\documentclass[twoside,10pt]{article}
\usepackage[accepted]{icml2015}
\usepackage{times}

\usepackage{amssymb}
\usepackage{amsmath}
\usepackage{mathrsfs}  
\usepackage{dsfont}
\usepackage{url}
\usepackage{tensor}
\usepackage{color}
\usepackage{natbib}
\usepackage{pgfplots}
\usepackage{psfrag}
\usepackage[off]{auto-pst-pdf}
\pgfplotsset{compat=newest}
\pgfplotsset{plot coordinates/math parser=false}
\usetikzlibrary{external}

\newcommand{\yb}{\mathbf{y}}

\newcommand{\zb}{\mathbf{z}}
\newcommand{\zbp}{\mathbf{z}^{\prime}}
\newcommand{\zbm}{\zb^{(m)}}

\newcommand{\zpr}{z^{\prime}_r}
\newcommand{\zr}{z_r}
\newcommand{\xb}{\mathbf{x}}
\newcommand{\xr}{x_r}
\newcommand{\xbi}{\xb^{(i)}}
\newcommand{\xbj}{\xb^{(j)}}

\newcommand{\xbn}{\xb^{(n)}}

\newcommand{\mb}{\mathbf{m}}

\newcommand{\ub}{\mathbf{u}}

\newcommand{\Kb}{\mathbf{K}}
\newcommand{\kb}{\mathbf{k}}

\newcommand{\Kuu}{\Kb_{zz}}
\newcommand{\Kxxp}{\Kb_{xx^\prime}}
\newcommand{\kxx}{\kb_{xx}}

\newcommand{\kxu}{\kb_{xz}}
\newcommand{\kux}{\kb_{zx}}

\newcommand{\kuxp}{\kb_{zx^\prime}}

\newcommand{\iKuu}{\Kuu^{-1}}
\newcommand{\Sb}{\mathbf{S}}
\newcommand{\Lb}{\mathbf{L}}

\newcommand{\Ex}{\mathbb{E}}
\newcommand{\No}{\mathcal{N}}
\newcommand{\Lo}{\mathcal{L}}

\newcommand{\Tr}{\mathrm{Tr}}
\newcommand{\KL}{\mathrm{KL}}

\newcommand{\Var}{\mathrm{Var}}
\newcommand{\vech}{\mathrm{vech}}

\newcommand{\dd}{\mathrm{d}}

\newcommand{\tr}{\mathrm{tr}}

\newcommand{\GP}{\mathcal{GP}}
\newcommand{\data}{\mathscr{D}}
\newcommand{\held}{\mathscr{H}}
\newcommand{\induce}{\mathcal{Z}}

\newcommand{\region}{\mathcal{T}}
\newcommand{\domain}{\mathcal{X}}

\newcommand\bigO{\mathcal{O}}

\newcommand{\ud}{\mathrm{d}}

\newcommand{\fx}{f_x}
\newcommand{\fn}{f_n}
\newcommand{\mux}{\tilde{\mu}_x}
\newcommand{\mun}{\tilde{\mu}_n}
\newcommand{\sigx}{\tilde{\sigma}_x}
\newcommand{\sign}{\tilde{\sigma}_n}

\newcommand\blfootnote[1]{%
  \begingroup
  \renewcommand\thefootnote{}\footnote{#1}%
  \addtocounter{footnote}{-1}%
  \endgroup
}

\newif\ifanonymised
\anonymisedfalse

\icmltitlerunning{Variational Inference for Gaussian Process Modulated Poisson Processes}

\usepackage{graphicx}
\begin{document}








\twocolumn[
\icmltitle{Variational Inference for Gaussian Process Modulated Poisson Processes}

\icmlauthor{Chris Lloyd*}{clloyd@robots.ox.ac.uk}
\icmlauthor{Tom Gunter*}{tgunter@robots.ox.ac.uk}
\icmlauthor{Michael A. Osborne}{mosb@robots.ox.ac.uk}
\icmlauthor{Stephen J. Roberts}{sjrob@robots.ox.ac.uk}
\icmladdress{Department of Engineering Science, University of Oxford}

\icmlkeywords{boring formatting information, machine learning, ICML}

\vskip 0.3in
]

\begin{abstract}
We present the first fully variational Bayesian inference scheme for continuous Gaussian-process-modulated Poisson processes. 
Such point processes are used in a variety of domains, including neuroscience, geo-statistics and astronomy, but their use is hindered by the computational cost of existing inference schemes.
Our scheme: requires no discretisation of the domain; scales linearly in the number of observed events; and is many orders of magnitude faster than previous sampling based approaches.
The resulting algorithm is shown to outperform standard methods on synthetic examples, coal mining disaster data and in the prediction of Malaria incidences in Kenya. 

\end{abstract}

\section{{Introduction}}
Sparse events defined over a continuous domain arise in a variety of real-world applications. 
The geospatial spread of disease through time, for example, may be viewed as a set of infections which occur in three dimensional space-time. 
In this work, we will consider data where the intensity (or average incidence rate) of the event generating process is assumed to vary smoothly over the domain. 
A popular model for such data is the \emph{inhomogenous Poisson process} with a \emph{Gaussian process model} for the smoothly-varying intensity function. 
This flexible approach has been adopted for applications in neuroscience \citep{Sahani*Byron*Cunningham*Shenoy*2007}, finance \citep{Basu*Dassios*2002} and forestry \citep{Heikkinen*Arjas*1999}.

However, existing inference schemes for such models scale poorly with the number of data, preventing them from finding greater use. 
The use of a full Gaussian process in modelling the intensity \citep{Adams_09_a} incurs prohibitive $\bigO(N^3)$ computational scaling in the number of data points, $N$.
To tackle this problem in practice, many approaches \citep{Rathbun*Cressie*1994,Moller_98_a} discretise the domain, binning counts within each segment. 
This approach enabled \citet{cunningham_fast_2008} to achieve $\bigO(N \log N)$ performance.
However, the discretisation approach suffers from poor scaling with the dimension of the domain and sensitivity to the choice of discretisation.

We introduce a new model for Gaussian-process-modulated Poisson processes that eliminates the requirement for discretisation, while simultaneously delivering $\bigO(N)$ scaling. 
We further introduce the first fully variational Bayesian inference scheme for such models, allowing computation many orders of magnitude faster than existing schemes.
This approach, which we term Variational Bayes for Point Processes (VBPP), is shown to provide more accurate prediction than benchmarks on held-out data from datasets including synthetic examples, coal mining disaster data and Malaria incidences in Kenya. 
The power of our approach suggests many future applications: in particular, our fully generative model will permit the joint inference of real-valued covariates (such as log-rainfall) and a point process (such as disease outbreaks).

\section{{Cox Processes}}
Formally\blfootnote{\bf{*} \textnormal{Corresponding authors.}} a Cox process---a particular type of inhomogenous Poisson process---is defined via a stochastic intensity function $\lambda(\xb) : \domain \rightarrow \mathbb{R}^+$.
For a domain $\mathcal{X} = \mathbb{R}^R$ of arbitrary dimension $R$, the number of points, $N(\mathcal{T})$, found in a subregion $\region \subset \mathcal{X}$ is Poisson distributed with parameter $\lambda_{\region} = \int_{\region} \lambda(\xb) \, \ud \xb$---where $\dd \xb$ indicates integration with respect to the Lebesgue measure over the domain---and for disjoint subsets $\region_i$ of $\domain$, the counts $N(\region_i)$ are independent.
This independence is due to the completely independent nature of points in a Poisson process \citep{Kingman_93_a}.

If we restrict our consideration to some bounded region, $\region$,  the probability density of a set of $N$ observed points, $\data = \{\xbn \in \region \}_{n=1}^N$, conditioned on the rate function $\lambda(\xb)$ is
\begin{equation}
p(\data \mid  \lambda ) = \exp \left\{- \int_{\region} \, \lambda(\xb) \,\dd \xb  \right\} \prod\limits_{n=1}^N \lambda(\xbn) \mathrm{.}
\label{eq:datalikelhood}
\end{equation}
We use $|\region|$ to denote the measure of the continuous domain $\region$.
In this work we will assume $\region$ is a hyper-rectangular sub-set of $\mathbb{R}^R$ with boundaries $\region_r^{\mathrm{min}}$ and  $\region_r^{\mathrm{max}}$ in each dimension $r$ and 
\begin{equation}
|\region| = \int_\region 1\; \dd \xb = \prod_{r=1}^R (\region_r^{\mathrm{max}} - \region_r^{\mathrm{min}}).
\end{equation}
Using Bayes' rule, the posterior distribution of the rate function conditioned on the data, $p(\lambda | \data)$, is
\begin{align}
\frac{p(\lambda)\exp \left\{- \int_{\region} \, \lambda(\xb) \,\dd \xb  \right\} \prod\nolimits_{n=1}^N \lambda(\xbn)}{\int p(\lambda)\exp \left\{- \int_{\region} \, \lambda(\xb) \,\dd \xb  \right\} \prod\nolimits_{n=1}^N \lambda(\xbn) \dd \lambda} \label{eqn:intracint}
\end{align}
which is often described as ``doubly-intractable'' because of the nested integral in the denominator.

\subsection{Inferring Intensity Functions}

To overcome the challenges posed by the doubly intractable integral \citet{Adams_09_a} propose the Sigmoidal Gaussian Cox Process (SGCP).
In the SGCP, a Gaussian process \citep{Rasmussen_06_a} is used to construct an intensity function prior by passing a random function, $f \sim \GP$, through a sigmoid transformation and scaling it with a maximum intensity $\lambda^*$.
The intensity function is therefore $\lambda(x) = \lambda^* \sigma \bigl(f(x)\bigr)$, where $\sigma(\cdot)$ is the logistic sigmoid (squashing) function
\begin{align}
\sigma(x) = \frac{1}{1+\exp(-x)}. 
\end{align}
To remove the inner intractable integral, the authors augment the variable set to include latent data, such that the joint distribution of the latent and observed data is uniform Poisson over the region $\mathcal{T}$.
While this model works well in practice on small, sparse event data in one dimension, in reality, it scales poorly with both the dimensionality of the domain and the maximum observed density of points.
This is due to: the incorporation of latent, or thinned, data, whose number grows exponentially with the dimensionality of the space; and an $\bigO(N^3)$ cost in the number $N$ of all data (thinned or otherwise).

In \citet{Gunter*Lloyd*Osborne*Roberts*2014}, the authors go some way towards improving the scalability of the SGCP, by introducing a further set of latent variables such that the entire space need no longer be thinned uniformly.
Instead, they thin to a piecewise uniform Poisson process, maintaining the tractability of the inner integral, and allowing the model to scale to higher dimensional point processes. The authors term this approach ``adaptive thinning''.

In \citet{Lasko14} the author performs renewal process inference without thinning the domain, by making use of a positively transformed intensity function.  The intractability of their chosen approach forces them to resort to numerical integration techniques, however, and Bayesian inference is still performed using computationally expensive sampling.

\section{{Model}}

We construct our prior over the rate function using a Gaussian process.
Rather than using a squashing function, we will assume\footnote{See Section \ref{sec:othertranforms} for a detailed motivation of this choice.} the intensity function is simply defined as $\lambda(\xb) = f^2(\xb), \; \xb \in \region$, where $f$ is a Gaussian process distributed random function achieving a non-negative prior \citep{gunter14-fast-bayesian-quadrature}.
Furthermore we will assume that $f$ is dependent on a set of \emph{inducing} points $\induce = \{\zbm \in \region\}_{m=1}^M$. We denote the evaluation of $f$ at these points $\ub$, and note $\ub$ has distribution $\ub \sim \No( \vec{1}\bar{u}, \Kuu )$.
Using this formulation $f|\ub \sim \GP \bigl(\mu(\xb), \Sigma(\xb,\xb^\prime)\bigr)$ has mean and covariance functions
\begin{align}
\mu(\xb)               &= \kxu \iKuu \ub, \\
\Sigma(\xb,\xb^\prime) &= \Kxxp - \kxu\iKuu \kuxp,
\end{align}
where $\kxu$, $\Kxxp$, $\Kuu$ are matrices evaluated at $\xb$, $\xb^\prime$ and $\induce$ using an appropriate kernel.
We use the exponentiated quadratic (also known as the ``squared exponential'') ARD kernel
\begin{equation}
\label{eq:ard_cov}
K( \xb, \xb' )\!=\!\gamma \prod_{r=1}^R \exp \left( -\frac{(x_r-x'_r)^2}{2\alpha_r} \right).
\end{equation}

With this hierarchical formulation the joint distribution over $\data$, $f$, $\ub$ and $\Theta$ is
\begin{align}
p(\data, f, \ub, \Theta )\!=\!p( \data | \lambda=f^2 ) p( f | \ub, \Theta ) p( \ub | \Theta )p(\Theta),
\end{align}
where $p(\Theta)$ is an optional prior on the set of model parameters $\Theta = \{ \gamma, \alpha_1, \hdots, \alpha_R, \bar{u} \}$. For notational convenience we will often omit conditioning on $\Theta$.

\section{{Inference}}

We will use variational inference to obtain a bound on the model evidence $p(\data)$.
To achieve this we must integrate out $f$ and $\ub$, but we must also integrate $f^2$ over the region $\region$ due to the integral embedded in the likelihood, Equation \ref{eq:datalikelhood}.

\subsection{Variational Bound}

We begin by integrating out the inducing variables $\ub$, using a variational distribution $q(\ub) = \No( \ub;\mb, \Sb )$ over the inducing points.
We now multiply and divide the joint by $q(\ub)$ and lower bound using Jensen's inequality to obtain a lower bound on the model evidence:
\begin{align}
\log p( \data|\Theta ) &= \log\left[ \int \hspace{-0.5em}\int p( \data | f ) p(f|\ub) p(\ub) \frac{q(\ub)}{q(\ub)} \; \dd \ub \; \dd f \right] \notag\\
&\ge   \int\hspace{-0.5em} \int p(f|\ub) q(\ub) \; \dd \ub \; \log[ p( \data | f )] \; \dd f \notag \\
&+   \int\hspace{-0.5em}\int p(f|\ub) q(\ub) \; \dd f \log\left[ \frac{p(\ub)}{q(\ub)}\right] \; \dd \ub \notag \\
&= \Ex_{q(f)} \left[ \log p( \data  | f ) \right] - \KL\bigl(q(\ub) ||p( \ub )\bigr) \notag \\
&\triangleq\Lo
\end{align}
Since  $p(f|\ub)$ is conjugate to $q(\ub)$, we can write down in closed-form the resulting integral:
\begin{align}
q( f ) &= \int p( f | \ub ) q( \ub ) \dd \ub = \GP( f; \tilde{\mu}, \tilde{\Sigma}  ), \label{eq:var_f}\\
\tilde{\mu}(\xb)               &= \kxu \iKuu \mb , \notag \\
\tilde{\Sigma}(\xb,\xb^\prime) &= \Kxxp - \kxu\iKuu \kuxp + \kxu\iKuu\Sb\iKuu\kuxp. \notag
\end{align}
$\KL \bigl(q(\ub) ||p( \ub )\bigr)$ is simply the KL-divergence between two Gaussians
\begin{align}
\KL \bigl(q(\ub) ||p( \ub )\bigr)
&= \frac{1}{2}\left[ \tr\left(\iKuu \Sb\right) - \log\frac{|\Kuu|}{|\Sb|} - M \right. \notag \\
&+\left. (\vec{1}\bar{u}-\mb)^\top\iKuu(\vec{1}\bar{u}-\mb)\right].
\end{align}
We can now take expectations of the data log-likelihood under $q(f)$: 
\begin{align}
\Lo &= \Ex_{q(f)} \left[ \log p( \data  | f ) \right] - \KL\bigl(q(\ub) ||p( \ub )\bigr) \notag \\
               &= \Ex_{q(f)} \left[ - \int_{\region} \, \fx^2 \,\dd \xb + \sum\limits_{n=1}^N \log \fn^2 \right] \notag \\
               &\phantom{=\ }- \KL \bigl(q(\ub) ||p( \ub )\bigr) \notag \\
               &= - \int_{\region} \left\{ \Ex_{q(f)}[\fx]^2 + \Var_{q(f)}[\fx] \right\}\dd \xb\notag \\
               &\phantom{=\ }+ \sum\limits_{n=1}^N \Ex_{q(f)} \left[ \log \fn^2 \right] - \KL \bigl(q(\ub) ||p( \ub )\bigr), \label{eq:lowerbound_start}
\end{align}

where to keep the notation concise we have introduced the following identities:
\begin{equation}
\begin{array}{ccc}
f_x \triangleq f(\xb),  & \mux \triangleq \tilde{\mu}(\xb),  & \sigx^2 \triangleq \tilde{\Sigma}(\xb,\xb), \\
f_n \triangleq f(\xbn), & \mun \triangleq \tilde{\mu}(\xbn), & \sign^2 \triangleq \tilde{\Sigma}(\xbn,\xbn).
\end{array}\notag
\end{equation}
Note we have used Tonelli's Theorem to reverse the ordering of the integrations over the positive integrand $f_x^2 q(f)$.
We now have two tasks remaining: we must compute the integral over the region $\region$ and calculate the expectations $\Ex_{q(f)} \left[ \log \fn^2 \right]$ at the data points.

\subsection{Integrating Over The Region $\region$}

This lower bound has the desirable property that we can take expectations under $q(f)$ at any specific point, $\xb$, of the function value, $f(\xb)$, since $q(f(\xb))$ is Gaussian.
It is only possible to take useful expectations because: a) we used the conditional GP formulation, allowing tractable expectations to be taken w.r.t. $q(f)$; b), we have already integrated out the inducing variables $\ub$; and c) we chose a suitable transformation, i.e. $\lambda(\xb)=f^2(\xb)$.

The required statistics for Equation \ref{eq:lowerbound_start} are:
\begin{align}
\Ex_{q(f)}[\fx]^2 = \mux^2    &= \mb^\top \iKuu \kux \kxu \iKuu \mb, \\
\Var_{q(f)}[\fx]  = \sigx^2 &= \kxx - \Tr( \iKuu \kux\kxu ) \notag \\
              &+ \Tr( \iKuu\Sb\iKuu \kux\kxu ) .
\end{align}
It is now easy to calculate the integral since only $\kux = \kxu^\top$ is a function of $\xb$, leading to the following terms:
\begin{align}
\int_{\region} \Ex_{q(f)}[\fx]^2 \dd \xb &= \mb^\top \iKuu \Psi \iKuu \mb, \\
\int_{\region} \Var_{q(f)}[\fx]  \dd \xb &= \gamma|\region| - \Tr( \iKuu \Psi ) \notag \\
                                      &+ \Tr(\iKuu\Sb\iKuu \Psi ) . 
\end{align}
For the exponentiated quadratic ARD kernel, the matrix 
\begin{equation}
\Psi = \int K(\zb,\xb)K(\xb,\zb^\prime) \; \dd \xb \label{eq:phi_stat}
\end{equation}
can be calculated by re-arranging the product as a single exponentiated quadratic in $\xb$ and $\bar{\zb}$
as follows:
\begin{align}
\Psi(\zb,\zbp) &= \int_{\region} \gamma^2\prod_{r=1}^R \exp \left( -\frac{( \zr - \zpr )^2}{4\alpha_r} -\frac{( \xr - \bar{z}_r )^2}{\alpha_r} \right) \dd \xb \notag \\
              &= \gamma^2\prod_{r=1}^R -\frac{\sqrt{\pi\alpha_r}}{2}\exp\left(-\frac{( \zr - \zpr )^2}{4\alpha_r}\right) \notag \\
              &\phantom{=\ }\times \left[ \mathrm{erf}\left(\frac{\bar{z}_r - \region_r^{\mathrm{max}})}{\sqrt{\alpha_r}} \right) -\mathrm{erf}\left(\frac{ \bar{z}_r - \region_r^{\mathrm{min}}}{\sqrt{\alpha_r}} \right)\right], \notag
\end{align}
where $\bar{\zb} = [\bar{z}_1,\hdots,\bar{z}_R]^\top$ has elements $\bar{z}_r = \frac{z_r + z_r^\prime}{2}.$

In addition to the exponentiated quadratic ARD kernel, the matrix $\Psi$ can be computed in closed-form for other kernels, including polynomial and periodic kernels, as well as sum and product combinations of kernels.

\subsection{Expectations At The Data Points}

The expectation $\Ex_{q(f)}[\log \fn^2]$ has an analytical---albeit complicated---solution expressed as
\begin{align}
\Ex_{q(f)}[\log \fn^2] &= \int\nolimits_{-\infty}^\infty \log (\fn^2) \; \No( \fn, \mun, \sign^2 ) \; \dd \fn\label{eq:signintegrate} \\
&= -\tilde{G}\left( -\frac{\mun^2}{2\sign^2} \right) + \log\left( \frac{\sign^2}{2} \right) - C, \label{eq:datapoint_expectations}
\end{align}
where $C\approx0.57721566$ is the Euler-–Mascheroni constant and $\tilde{G}$ is defined via the confluent hyper-geometric function 
\begin{align}
\tensor[_1]{F}{_1}(a,b,z) = \sum_{k=0}^{\infty} \frac{(a)_k z^k}{(b)_k \; k!},
\end{align}
where $(\cdot)_k$ denotes the rising Pochhammer series
\begin{equation}
\begin{array}{ll}
(a)_0 = 1, & (a)_k = a(a+1)(a+2)\hdots(a+k-1). \notag
\end{array}
\end{equation}
Specifically $\tilde{G}$ is a specialised version of the partial derivative of $\;\tensor[_1]{F}{_1}$ with respect to its first argument and can be computed using the method of \citet{Ancarani_08_a}, which has a particular solution at $a=0$, leading to the following definition of $\tilde{G}$:
\begin{align}
\tilde{G}(z) = \tensor[_1]{F}{_1}^{(1,0,0)}\left(0,\frac{1}{2},z\right) = 2z \sum_{j=0}^\infty \frac{j! \; z^j}{(2)_j (1\frac{1}{2})_j } \label{eq:G_tilde}.
\end{align}

\begin{figure}[htp]
\centering
\includegraphics[width=\columnwidth, trim = 9mm 0mm 10mm 4mm, clip]{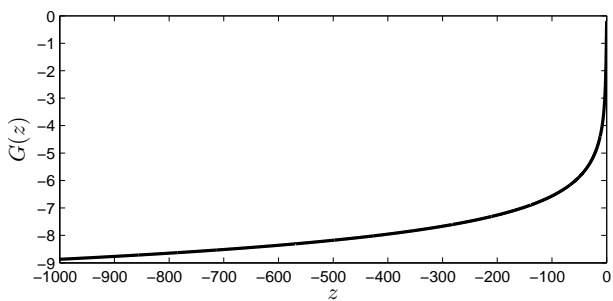}
\caption{Accurate evaluations of $\tilde{G}$ can be obtained from a precomputed look-up table.}
\label{fig:Gtilde}
\end{figure}

Naive implementation of Equation \ref{eq:G_tilde} has poor numerical stability, although this can be improved somewhat using an iterative scheme, in practice we therefore use a large multi-resolution look-up table of precomputed values obtained from a numerical-package.
As shown in Figure \ref{fig:Gtilde}, this function decreases very slowly as its argument becomes increasingly negative, so we can easily compute accurate evaluations of $\tilde{G}(z)$ for any $z$ by linear interpolation of our lookup table and, as a by-product, we also obtain $\tilde{G}^\prime$.


\vspace{-1mm}
\subsection{Optimising The Bound}
To perform inference we find variational parameters $\mb^*$, $\Sb^*$ and the model parameters $\Theta^*$ that maximise $\Lo$.
To optimise these simultaneously we construct an augmented vector $\yb = [ \Theta^\top, \mb^\top, \vech(\Lb)^\top]^\top$---where $\vech(\Lb)$ is the vectorisation of the lower triangular elements of $\Lb$, such that $\Sb=\Lb\Lb^\top$.

As well as the  maximum-likelihood solution, we can also compute the the maximum-a-posterior (MAP) estimate by maximising $\Lo(\data;\yb)+\log p(\Theta)$.

To optimise the inducing point locations we use the change of variables
\begin{equation}
z_r^{(m)} = \frac{\region_r^{\mathrm{min}}+\region_r^{\mathrm{max}}}{2} - \frac{\region_r^{\mathrm{min}}-\region_r^{\mathrm{max}}}{2} \sin\bigl( \omega_r^{(m)} \bigr) ,
\end{equation}
and optimise in $\omega_r^{(m)} \in [-\pi,\pi]$, which ensures the inducing points always remain within the region $\region$. 

\subsection{Locating The Inducing Points}

One final part of our model we have so far left unspecified is the number and location of inducing points.
In principle, for a given set of parameters $\Theta$, we will obtain a lower bound for the true GP likelihood for any number of inducing points in any configuration of locations. 
We consider two possible approaches: firstly, treating the inducing points as optimisation parameters and, secondly, fixing them on a regular grid.

If the locations of the inducing points are optimised, this suggests---in common with other sparse GP models---that we can achieve good performance using a small number of well-placed inducing points.
This is achieved at the cost of an additional set of optimisation parameters, although the size of $\mb$ and $\Sb$ are commensurately reduced.
Optimisation of the inducing points is particularly computationally expensive, however, because of the necessity to recompute $\iKuu$ for each dimension of each inducing point ($M \times R$ in total) and since $\Kuu$ affects every term in the bound.

Regular grids, on the other hand, also have several advantages. Consider firstly that---in contrast to standard sparse GP regression---the accuracy of our solution is not only governed by the distance between the inducing points and the data points.
The variance of $f(\xb)$ increases as $\xb$ becomes further from the inducing points.
However, the rate function, $\lambda$, is a function of both the mean \emph{and} the variance of $f$.
Since we are integrating $\lambda$ over $\region$, we need inducing points distributed across $\region$ and not just in regions close to the data.
An evenly-spaced grid is one way to ensure this is achieved.

Regularly sampled grids also afford potential computational advantages.
When the grid points are evenly spaced, the kernel matrix has Toeplitz structure, and hence allows matrix inversion (and linear solving) in $\bigO(M \log^2 M)$ time, a fact previously utilised for efficient point processes by \citet{cunningham_fast_2008}.
Furthermore, when the kernel function is separable across the dimensions (as specified by Equation \eqref{eq:ard_cov}), the kernel matrix has Kronecker structure which can further reduce the cost of matrix inversion \citep{osborne_real-time_2012}.
The latter is relevant to all sparse GP applications based on inducing points, however, it is particularly relevant for this application as we are motivated by the doubly intractable nature of Equation \ref{eqn:intracint}.
In our implementation, we use na\"{i}ve inversion of the inducing point kernel matrix, $\Kuu$, resulting in computational complexity of $\bigO(NM^3)$.
Hence the computation times reported below could be improved with a relatively small amount of additional implementation effort.  


\subsection{Predictive Distribution}\label{sec:predictive}

To form the predictive distribution we assume our optimised variational distribution $q^*(\ub) = \No(\ub; \mb^*, \Sb^*)$ approximates the posterior $p(\ub | \data)$. 
Analogously to Equation \ref{eq:var_f} we next compute $q^*( f ) \approx p( f | \data )$.
We can now derive a lower bound of the (approximate) predictive log-likelihood on a held-out test dataset $\held$:
\begin{align}
	\log p(\held | \data, \Theta^* ) &= \log \Ex_{p(f|\data)}[ p( \held | f ) ] \notag \\
	                                 &\approx \log \Ex_{q^*(f)}[ p( \held | f ) ] \triangleq \mathcal{M}_p  \notag \\
	                                 &\ge \Ex_{q^*(f)}[ \log p( \held | f ) ] \triangleq \Lo_p.
\end{align}
The derivation of $\Lo_p$ follows Equations \ref{eq:lowerbound_start}-\ref{eq:datapoint_expectations}.
The resulting bound is the same as $\Lo$ except $\mb$, $\Sb$ are replaced with $\mb^*$ and $\Sb^*$, and there is no KL divergence term. Kernel matrices are computed using $\Theta^*$.

The tightness of this final bounding step will be a function of how well the inducing variables $\ub$ define the function $f$.
Intuitively, when the variance at the inducing points is large, the entropy of $f$ will be large, as it is unrestricted.
From an information theoretic perspective, we can say that the tightness of the bound will be a function of the entropy of $f$: $H(f) = \frac{1}{2}\log | \tilde{\Sigma} | + \text{const}$.

Given this knowledge we define a second, tightened, lower bound $\Lo_0$, where we allow the variance of function values at the inducing points to collapse to zero: $p(\ub|\data) \approx \delta(\mb)$.
This reduces the conditional entropy of $f$ given $\ub$ by shrinking the final term in the definition of $\tilde{\Sigma}$ (Equation \ref{eq:var_f}), resulting in a tighter bound for a slightly restricted class of models.
As it is a slightly more restricted model, we expect the ground truth $\mathcal{M}_0 = \log \left(\mathbb{E}_{q(f|\ub=\mb)}\left[p(\mathcal{H}|f) \right]\right)$ to be lower than the ground truth $\mathcal{M}_p$.
In practice, however, because the variational $\Lo_0$ is so much tighter, we also use this to give results for approximate predictive likelihood when comparing against other approaches.
In Figure \ref{fig:predictive_bound} we demonstrate this empirically for the coal mining dataset by evaluating the true predictive bounds on a held-out $50\%$ of the data via $10,000$ MCMC samples, and shading between these and the variational approximations.
We do this for a range of inducing point grid densities, both with and without optimisation of the inducing point locations. In Figures \ref{fig:twitter_Z_opt}, \ref{fig:twitter_grid}, and \ref{fig:predictive_bound}, we plot all of the bounds described above as the number of inducing points increase.
We note that for the relatively smooth coal mining data, (Figure \ref{fig:predictive_bound}), all bounds do not benefit from more than about $10$ inducing points, while in the case of the twitter data, the faster dynamics call for increased numbers.
In all Figures the tightness of $\Lo_0$ is evident as compared to $\Lo_p$. 

\section{{Alternative GP Transformations}}\label{sec:othertranforms}

At this point it is worth considering why we have chosen the function transformation $ \lambda(\xb) = f^2(\xb)$ in preference to other alternatives we might have used.
An obvious first choice would be 
\begin{equation}
\lambda(\xb) = \exp\bigl(f(\xb)\bigr). 
\end{equation}
This transformation is undesirable for two reasons.
The more obvious of these is that after taking expectations under $q(f)$ we are left with the integral
\begin{equation}
-\int_\region \exp\left( \mux + \frac{\sigx^2}{2} \right) \dd \xb,
\end{equation}
which cannot be computed in closed form. We could approximate the integral using a series expansion, however this would be difficult with more than a couple of terms and furthermore, since the function is concave, this approximation would not be a lower bound.

The second---and more subtle---reason is that in using this transformation, when we take expectations under $q(f)$ of the data, we obtain
\begin{align}
\Ex_{q(f)} \left[ \log \left\{ \exp( \fn ) \right\} \right] = \mun.
\end{align}
Since the mean, $\mun$, is not a function of $\Sb$, the variance of the variational distribution $q(\ub)$, we have effectively decoupled the data from the uncertainty on our variational approximation; this is clearly undesirable.

Another possible candidate is the probit function, $\lambda(x) = \Phi(f(x))$.
This can be integrated analytically against the GP prior, however we are again left with a difficult integral over $\region$ which is
\begin{align}
-\int_\region \Phi\left( \frac{\mux}{\sqrt{1+\sigx^2}} \right) \dd \xb .
\end{align}
As the range of this transformation is $(0,1)$ we would also require additional machinery to infer a scaling variable.

In contrast the square transform presented allows the integral over the region $\region$ to be computed in closed form and $\Ex_{q(f)}[\log \fn^2]$ can be computed quickly and accurately.
Importantly this transformation also maintains the connection between the data and the variational uncertainty. 

Although the square transform is not a one-to-one function---any rate function $\lambda$ may have been generated by $f^2$ or $(-f)^2$---this sign ambiguity is integrated out in a Bayesian sense, Equation \ref{eq:signintegrate}.

\section{{Relationship to Sparse GP Models}}

The use of inducing points in our model relates it to a wide range of sparse Gaussian process models, e.g. SPGP \cite{Snelson_05_a}.
The variational sparse Gaussian process framework was introduced by \citet{Titsias_09_a}, however the bound we develop is more akin to the ``Big-Data'' GP bound \cite{Hensman_13_a}, since we explicitly maintain the variational distribution $q(\ub)$.
The variable $\Psi$ that results from integrating the kernel over the input domain is similar to the so-called ``$\Psi$-statistic'' which arises when integrating out the uncertainty of latent variables in the variational GPLVM \cite{Titsias_04_a}.

\section{{Experiments}}
To evaluate our algorithm, we benchmarked against our frequentist kernel smoothing approach, described below, both with without end correction (KS+EC and KS--EC respectively), and a fully Bayesian SGCP MCMC sampler.
Our test data sets are generative data from the SGCP model and several real-world data sets.



\subsection{Benchmarks}

Our kernel smoothing method is similar to standard kernel density estimation except we use truncated normal kernels to account for our explicit knowledge of the domain---the latter is referred to as ``end-correction'' in some literature \citep{Diggle_85_a}.
The kernel smoother optimises a diagonal covariance, $\Sigma^*$, by maximising the leave-one-out training objective
\begin{align}
\Sigma^* = \underset{\Sigma}{\operatorname{argmax}} \sum_{i=1}^N \log \sum_{j\neq i=1}^N \No_\region(\xbi;\xbj,\Sigma) .
\end{align}
We can construct the predictive distribution by combining the maximum-likelihood estimates of the size and spatial location of the data.
For a test data set $\held$ (with $K!$ permutations) this distribution is
\begin{align}
p( \held | \data ) =  K! \; p( K | \data ) \prod_{k=1}^K p( \tilde{\xb}^{(k)} | \data ), \label{eq:kdelikelihood}
\end{align}
where the location density,
\begin{align}
p( \tilde{\xb}^{(k)} | \data ) = \frac{1}{N}\sum_{n=1}^N \No_\region( \tilde{\xb}^{(k)}; \xbn, \Sigma^* ), 
\end{align}
is computed using using the previously described method and the distribution of the number of points,
\begin{align}
p( K | \data ) = \frac{N^K}{K!}\exp( -N ),
\end{align}
is simply a Poisson distribution with parameter $N$.
It is straight forward to show that Equation \ref{eq:kdelikelihood} is equivalent to Equation \ref{eq:datalikelhood} since we can interpret the rate function as $\lambda(\xb) = \sum\nolimits_{n=1}^N \No_\region( \xb; \xbn, \Sigma^* )$ and since $ \int_{\region} \lambda(\xb) \dd \xb = N$.

Our SGCP sampler is based on \citet{Adams_09_a}.
Our implementation differs by using elliptical slice sampling to infer the latent function $f$ and we perform hyper-parameter inference using Hybrid Monte-Carlo (HMC). 
We also use the ``adaptive-thinning'' method described in \citet{Gunter*Lloyd*Osborne*Roberts*2014}, to reduce the number of thinning points required.
\subsection{Synthetic Data}
\begin{center}
\begin{table*}[t]
\center
\caption{Results for 2D synthetic data (drawn from the SGCP generative process).}
\begin{tabular}{cccccccccc}
\multicolumn{1}{c}{} & \multicolumn{3}{c}{{\bf{SGCP}}} & \multicolumn{3}{c}{{\bf{KS+EC}}} & \multicolumn{3}{c}{{\bf{VBPP} ($\Lo_0$)}} \\ 
 {Function}  & Avg. LL & RMS  & Time(s) &  Avg. LL & RMS  & Time(s) & Avg. LL & RMS  & Time(s) \\  \hline
 1  & 446.1 & 1.37 & 7547.83 & 389.8 & 1.48 & 0.34 & 392.9 & 1.21 & 3.26 \\ 
 2  & -61.1 & 0.38 & 1039.65 & -78.3 & 0.46 & 0.02 & -76.1 & 0.38 & 2.00 \\ 
 3  & 122.4 & 0.88 & 3173.91 & 84.3  & 1.04 & 0.12 & 92.6  & 0.81 & 2.44 \\ 
 4  & 175.8 & 1.71 & 3773.75 & 147.0 & 1.26 & 0.05 & 148.3 & 1.14 & 2.58 \\ 
 5  & 446.1 & 2.94 & 6368.44 & 413.6 & 2.02 & 0.21 & 415.5 & 1.81 & 2.83 \\
\end{tabular}\label{table:2dgenlog}\end{table*}

\end{center}
\begin{figure}[htp]
\centering
\includegraphics[width=0.9\columnwidth, trim = 0mm 2mm 0mm 2mm, clip]{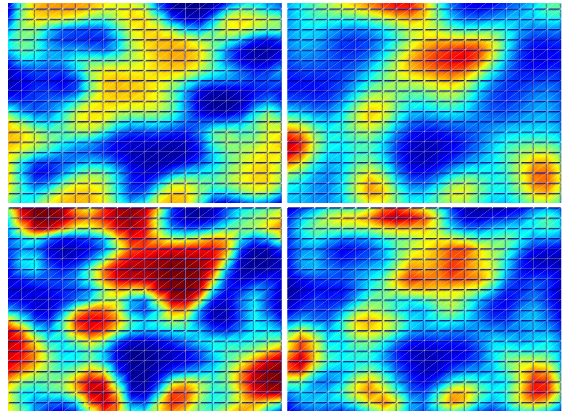}
\caption{\centering{2D Synthetic Data. Clockwise from top left: Ground truth, VBPP, KS+EC, SGCP.}}\label{fig:gen2d}
\end{figure}
For the synthetic data sets, we first generate a 2D function from a high resolution grid using a Gaussian process and sigmoid link function, and then, conditioned on that function, we draw a training dataset and multiple test datasets.
We give average performances results for these test datasets in Table \ref{table:2dgenlog}.
Figure \ref{fig:gen2d} visualises an inferred 2D intensity conditioned on $\sim 500$ observations.
Although the SGCP sampler gives better predictive performance than the VBPP $\Lo_0$ bound, it should be noted that the sampler uses well tuned hyper-parameters, uses the same link function as the generative process and is much more computationally expensive.
VBPP outperforms kernel smoothing in terms of both predictive likelihood and RMS error.


\begin{figure*}[tp]
\centering
\begin{minipage}{0.42\textwidth}
\centering
\includegraphics[width=1.0\columnwidth, trim = 10mm 8mm 15mm 2mm, clip]{images/coalMining3_3rate_15_2_3_00_31.eps}
\caption{Coal mining disaster data.}
\label{fig:coal}
\end{minipage}
\begin{minipage}{0.14\textwidth}
\centering                     
\includegraphics[width=0.8\columnwidth, trim = 10mm 6mm 8mm 5mm, clip]{images/legend.eps} 
\end{minipage}
\begin{minipage}{0.42\textwidth}
\centering
\includegraphics[width=1.0\columnwidth, trim = 8mm 8mm 13mm 7mm, clip]{images/twitter32.eps}
\caption{\centering{Twitter data}}\label{fig:twitter}
\end{minipage}
\begin{minipage}{0.42\textwidth}
\centering
\includegraphics[width=1.0\columnwidth, trim = 9mm 6mm 15mm 2mm, clip]{images/twitter_Grid.eps}
\caption{Twitter data (Z on a fixed grid).}
\label{fig:twitter_grid}
\end{minipage}
\begin{minipage}{0.14\textwidth}
\centering                    
\includegraphics[width=0.8\columnwidth, trim = 6mm 4mm 6mm 4mm, clip]{images/legend_2.eps}
\end{minipage}
\begin{minipage}{0.42\textwidth}
\centering
\includegraphics[width=1.0\columnwidth, trim = 9mm 6mm 13mm 7mm, clip]{images/twitter_Z_opt.eps}
\caption{\centering{Twitter data (Z optimised).}}\label{fig:twitter_Z_opt}
\end{minipage}
\end{figure*}

\subsection{Real Data}

We next investigate three real world data sets.
For these data sets we create training and test subsets by allocating each point to either subset with probability 0.5.
Since true rates are unknown for these datasets we rely on held-out predictive likelihood as the only performance metric.

\subsubsection{Coal Mining Disaster Data} \label{sec:coal}
\begin{figure}[h]
\centering
\includegraphics[width=1.0\columnwidth, trim = 10mm 5mm 15mm 2mm, clip]{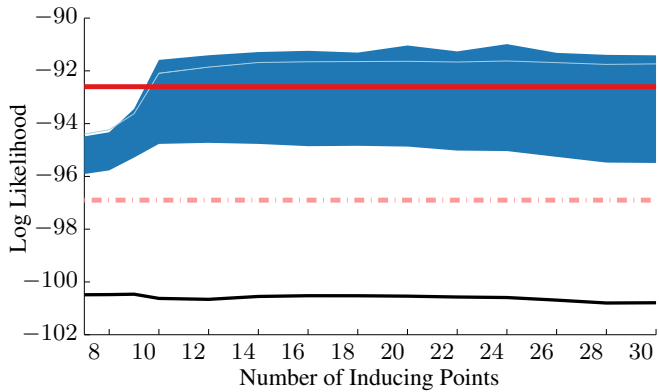}
\caption{Coal Mining Data set: The difference between sampling $\mathcal{M}_0$ and $\mathcal{M}_p$, and the corresponding lower bounds, $\Lo_0$ and $\Lo_p$ (see Section \ref{sec:predictive}). The figure clearly demonstrates the tightness of the $\Lo_0$ bound. Legend as in Figure \ref{fig:twitter_grid}.}\label{fig:predictive_bound} 
\end{figure}
Our first real dataset comprises 190 events recorded from March
15, 1851 to March 22, 1962; each represents a coal-mining disaster
that killed at least ten people in the United Kingdom. 
These data, first analysed in this form in $1979$
\citep{Jarrett_1979}, have often been tackled with nonhomogeneous Poisson processes, \citep{Adams_09_a}, as the rate of such disasters is expected to vary according to known historical developments.
The events are indicated by the rug plot along the axis of
Figure \ref{fig:coal}.

Our inferred intensity of disasters correlates with the historical introduction of safety regulation, as noted in previous work on this data \citep{Fearnhead_2006,Carlin*Gelfand*Smith_1992}.
Firstly, our results depict a decline in the rate of such disasters throughout 1870--1890, a period that saw the UK parliament passing several acts with the aim of improving safety for mine workers, including the Coal Mines Regulation Acts of 1872 and 1887.
Our inferred intensity also declines after 1950, likely related to the imposition of further safety regulation with the Mines and Quarries Act, 1954. 

Predictive log-likelihood values on held out data (Figure \ref{fig:predictive_bound}) are also encouraging.
VBPP outperforms kernel smoothing and SGCP with as few as 10 inducing points; more inducing points yielding no further benefit.
\subsubsection{Twitter Data}
Next, we ran the models on the tweet profile of the chairman of the `Better Together Campaign', Alistair Darling, one week either side of the Scottish independence election (189 tweets).
Results are shown in Figure \ref{fig:twitter} and Table \ref{table:twitter}, where half the data was held out and a regular 31 point grid was used.
Figures \ref{fig:twitter_grid} and \ref{fig:twitter_Z_opt} compare the performance of regularly spaced and optimised inducing points, and show optimisation yields considerably improved performance on this dataset. The $\Lo_0$ and $\Lo_p$ bounds become less tight as the number of inducing points is increased, suggesting there is less uncertainty represented in the variational parameter $\Sb$ and more uncertainty captured by a reduced kernel length scale.
This transition is observed for fewer inducing points when inducing point optimisation is employed.
Both with and without inducing point optimisation, VBPP $\mathcal{M}_0$ and $\mathcal{M}_p$ outperform both the SGCP and kernel smoothing by a wide margin, suggesting the square link function is an appropriate model for this data.

\subsubsection{Malaria Data}
\begin{figure}[t]
\centering
\includegraphics[width=0.75\columnwidth, trim = 162mm 32mm 143mm 42mm, clip]{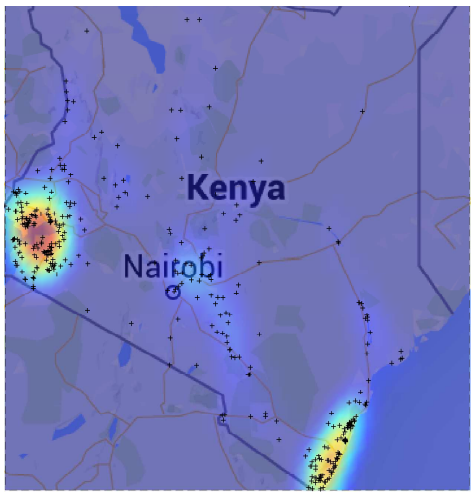}
\caption{A sample of $741$ malaria incidences in Kenya, which occured over the course of $1985$-$2010$, and the associated VBPP intensity function. $20\times20$ inducing points.}\label{fig:kenya}
\end{figure}
We expect that a major application of the contributions presented in this paper is the joint modelling of disease incidence with correlating factors, in a fully Bayesian, scalable framework.
For example, those studying the spread of malaria often wish to use continuous rainfall measurements to better inform their epidemiological models.
We use examples from the \citet{map} to test our scheme.
We extracted $741$ incidences of malaria outbreak documented in Kenya between $1985$ and $2010$, and ran our VBPP algorithm and kernel smoothing on approximately half of the resulting dataset, holding out the remainder for testing.
Test log-likelihood results, given in Table \ref{table:mos}, show VBPP performs comparably to kernel smoothing.
\begin{table}[htp]
\center
 \caption{Run times for 1D data sets.} 
 \begin{tabular}{ccc}
 \bf{\small{Method}} & \bf{\small{Coal Mining}}  & \bf{Twitter} \\ \hline
 \small{VBPP} & \small{0.7} & \small{0.5} \\ 
 \small{KS+EC}  & \small{0.0} & \small{0.3} \\ 
 \small{KS-EC}  & \small{0.0} & \small{0.2} \\ 
 \small{SGCP}  & \small{417.6} & \small{230.0} \\
\end{tabular}\label{table:twitter}
\caption{Test log-likelihood for 2D Malaria data.}
\begin{tabular}{cccc}
\multicolumn{1}{c}{\bf{\small{KS-EC}}} & \multicolumn{1}{c}{\bf{\small{KS+EC}}} & \multicolumn{1}{c}{\bf{\small{VBPP($\mathcal{M}_p$)}}} & \multicolumn{1}{c}{\bf{\small{VBPP($\Lo_0$)}}} \\ 
 \hline
\small{855.0} & \small{867.2} & \small{869.7} & \small{855.9}
\end{tabular}\label{table:mos}
\end{table}

\section{Further Work}
Although the performance of the variational Bayesian point process inference algorithm described in this paper improves upon standard methods when used in isolation, it is in its extensions that its utility will be fully realised.
Previous work \citep{Gunter*Lloyd*Osborne*Roberts*2014} has shown that hierarchical modelling of point processes---structured point processes---can significantly improve predictive accuracy.
In these multi-output models, statistical strength is shared across multiple rate processes via shared latent processes.
The method presented here provides a likelihood model for point-process data that can be incorporated as a probabilistic building-block into these larger interconnected models.
That is, our fully generative model can readily be extended to additionally incorporate other observation modalities. 
For example, real-valued observations such as (log-) household income could be modelled along with the intensity function over crime incidents using a variational multi-output GP framework.
Future work will be aimed towards developing these variational structured point process algorithms and integration of these techniques to popular Gaussian process tool-kits, such as GPy \citep{GPy2014}.

\section{{Conclusion}}

Point process models have hitherto been hindered by their scaling with the number of data.
To address this problem, we propose a new model, accommodating non-discretised intensity functions, that permits linear scaling. 
We additionally contribute a variational Bayesian inference scheme that delivers rapid and accurate prediction.
The scheme is validated on real datasets including the canonical coal mining disaster data set and malaria incidence in Kenya data.

\ifanonymised
\else
\subsubsection*{Acknowledgements}
The authors would like to thank James Hensman and Neil Lawrence for  helpful discussions.
Chris Lloyd is funded by a DSTL National PhD Scheme Studentship.
Tom Gunter is supported by UK Research Councils. 
\fi

\bibliographystyle{icml2015}
\bibliography{paper}






\end{document}